\def\year{2018}\relax
\documentclass[letterpaper]{article} %DO NOT CHANGE THIS
\usepackage{aaai18}  %Required
\usepackage{times}  %Required
\usepackage{helvet}  %Required
\usepackage{courier}  %RequiredAAAI
\usepackage{multirow}
\usepackage{booktabs}
\usepackage{amssymb}
\usepackage{amsmath}
\usepackage[table]{xcolor}
\usepackage{amsmath}
\usepackage{latexsym}
\usepackage{caption}
\usepackage{graphicx}
\usepackage{enumerate}
\usepackage{float}
\usepackage{url}  %Required
\usepackage{graphicx}  %Required
\begin{document}
\title{Learning Multimodal Word Representation via Dynamic Fusion Methods}
\author{Shaonan Wang$^{1,2}$, Jiajun Zhang$^{1,2}$, Chengqing Zong$^{1,2,3}$ \\
	$^1$ National Laboratory of Pattern Recognition, CASIA, Beijing, China \\  
	$^2$ University of Chinese Academy of Sciences, Beijing, China \\
	$^3$ CAS Center for Excellence in Brain Science and Intelligence Technology, Shanghai, China  \\
	\{shaonan.wang,jjzhang,cqzong\}@nlpr.ia.ac.cn}
\maketitle
\begin{abstract}
	Multimodal models have been proven to outperform text-based models on learning semantic word representations. Almost all previous multimodal models typically treat the representations from different modalities equally. However, it is obvious that information from different modalities contributes differently to the meaning of words. This motivates us to build a multimodal model that can dynamically fuse the semantic representations from different modalities according to different types of words. To that end, we propose three novel dynamic fusion methods to assign importance weights to each modality, in which weights are learned under the weak supervision of word association pairs. The extensive experiments have demonstrated that the proposed methods outperform strong unimodal baselines and state-of-the-art multimodal models.
\end{abstract}

\section{Introduction}

Representing the meaning of a word is a prerequisite to solve many natural language problems, such as calculating semantic relations between different words, finding the most relevant images of a word and so on. In recent years, computational semantic models that represent word meanings from patterns of word co-occurrence in corpora have received a lot of research interests \cite{turney2010frequency,mikolov2013efficient,clark2015vector}. However, compared to human semantic representation, these purely text-based models are severely impoverished for lacking perceptual information attached to the physical world. This observation has led to the development of multimodal word representation models that utilize both linguistic (e.g., text) and perceptual information (e.g., images, audios). Such models can learn better semantic word representations than text-based models, as evidenced by a range of evaluations \cite{andrews2009integrating,bruni2014multimodal,silberer2016visually}.

Learning good multimodal word representations relies not only on the quality of the word representations from linguistic and perceptual inputs, but also the ability to productively combine these representations. However, the existing multimodal models generally treat the word representations from different modalities equally. This is inconsistent with the fact that meaning of concrete words like \texttt{horse} and \texttt{computer} are mostly learned from perceptual experiences of seeing, touching and listening. In contrast, more abstract words, such as \texttt{hope} and \texttt{lovely}, are encoded mostly in linguistic modality rather than perceptual modality, which has been found in cognitive psychology \cite{wang2010neural,binder2016toward} and computational experiments \cite{hill2014multi,hill2014learning}. 

All these factors motivate us to build a multimodal model that can dynamically fuse information from linguistic and perceptual modalities according to different types of words. We can optimize the importance weights of different modalities for a word if the word has the gold representation. As no gold word representation exists in reality, we resort to word pairs which share the same meaning, so that they can guide each other. In this paper we utilize word association pairs\footnote{We have also tried other resources, such as synonyms from WordNet. However, these datasets are noisy and perform slightly worse, thus we only report results of word associations.}, which are generated by subjects firstly reading a cue word and then writing down the first word(s) that come to mind. Some examples are \texttt{wealthy} and \texttt{rich}, \texttt{jigsaw} and \texttt{puzzle}, \texttt{larger} and \texttt{bigger}. We assume that these association word pairs can lead us to learn the importance weights for different modalities. For instance, representations of abstract words \texttt{larger} and \texttt{bigger} are composed by linguistic and perceptual vectors, and the linguistic vectors are more important in representing abstract word meaning (i.e., the two words share more similarity in linguistic modality). To achieve the goal of making these two association words obtain similar representations, the model will assign more weights to their linguistic vectors. 

In light of these considerations, we propose three novel dynamic fusion methods to improve multimodal word representations. The three methods utilize a modality-specific gate, category-specific gate and sample-specific gate respectively, to learn different weights of linguistic and perceptual representations for each input modality, each supersense category and each word sample respectively. Furthermore, we perform extensive quantitative and qualitative analysis to shed light on the principle of the proposed dynamic fusion methods.
To summarize, our main contributions are two-fold:

\begin{itemize} 
	\item	We present a novel dynamic fusion method for multimodal representations, which utilizes a small set of word association pairs to learn different weights of different modalities for semantic word  representations. The core idea is to introduce weak supervision to learn a generic fusion rule. Results on six standard benchmarks demonstrate that our method significantly improves the quality of baseline multimodal representations.
	
	\item	Quantitative analysis shows that the proposed models can successfully assign different weights to linguistic and perceptual representations, and the learned weights show clear difference between concrete and abstract words. This offers initial support for the idea that humans differently encode concrete words and abstract words, and it also indicates that computational models can assist in exploring human semantic representation.
\end{itemize}

\section{Background and Related Work}
\subsection{Cognitive Grounding}

Dual coding theory \cite{hiscock1974imagery} posits that concrete words are represented in the brain in terms of a visual and linguistic code, whereas abstract words are encoded only in the linguistic modality. This theory has been initially validated by a number of neuroimaging studies \cite{wang2010neural,Anderson2017visual}.

In summary of previous studies, Wang et al. \shortcite{wang2010neural} conduct a meta-analysis for differences in human neural representation of abstract and concrete words. Their results show that abstract words elicit greater activity in linguistic-related brain area while concrete words elicit greater activity in perceptual-related brain area. With the help of computational models, Andrew et al. \shortcite{Anderson2017visual} decode functional Magnetic Resonance Imaging (fMRI) activity patterns associated with concrete and abstract words. They observe that both linguistic and visual representations can significantly decode most concrete nouns, while the abstract nouns can only be decoded by linguistic representations. 

To sum up, these studies hold that both linguistic and perceptual information affect human representations of concrete words, while only linguistic modality plays a large role in representing meaning of abstract words. In this respect, our method employs a representation process analogous to that of humans, in which linguistic and perceptual modalities contribute differently to concrete and abstract words.

%Theories of semantic representation assumes that concepts in the human brain can be represented by sets of features or attributes that are in some sense primitive or basic components of meaning. Based on brain-imaging datasets, \cite{binder2016toward} propose a componential model of semantic representations based entirely on functional divisions in the human brain. Specifically, they represent concepts by sets of physical and abstract properties like \textit{vision, somatic, audition, social, emotion, attention}, and provide normative data on the salience of each attribute for a large set of English words. Analysis show that abstract words are rated higher than concrete words on abstract properties (which are more encoded in linguistic modality) like \textit{emotion, social and attention}, and concrete words are rated higher on physical properties like \textit{vision, somatic, audition}. 

\subsection{Multimodal Models}
There is by now a large literature of multimodal representation models, and the existing models can be generally classified into two groups: 

\begin{enumerate}[(1)]
	\item \textbf{Joint training models} that build multimodal representations with raw inputs of both linguistic and perceptual resources.
	
	A class of models extends Latent Dirichlet Allocation \cite{blei2003latent} to jointly learn topic distributions from words and perceptual units \cite{fellbaum1998wordnet,andrews2009integrating,silberer2012grounded,roller2013multimodal}. The recently introduced work is an extension of the Skip-gram model \cite{mikolov2013efficient}. For instance, Hill and Korhonen \shortcite{hill2014learning} propose a corpus fusion method that inserts the perceptual features of a word in the training corpus, which is then used to train the Skip-gram model. Lazaridou et al. \shortcite{lazaridou2015combining} propose MMSkip model, which injects visual information in the process of learning linguistic representations by adding a max-margin objective function to minimize the distance between linguistic vectors and visual vectors.	
	
	The joint training methods implicitly propagate perceptual information to word representations and at the same time learn multimodal representations. However, these methods utilize raw text corpus in which words associated with perceptual information account for a small portion. This weakens the effect of introducing perceptual information, and consequently leads to only limited improvement of linguistic vectors.

	\item \textbf{Separate training models} that independently learn linguistic and perceptual representations and integrate them afterwards.
	
	The simplest approach is concatenation, which fuses linguistic and visual vectors by concatenating them. It has been proven to be effective in learning multimodal models \cite{bruni2014multimodal,hill2014multi,collell2017imagined}. Variations of this method employ transformation and dimension reduction on the concatenation result, including application of singular value decomposition (SVD) \cite{bruni2014multimodal} or canonical correlation analysis (CCA) \cite{hill2014multi}. In addition, Silberer and Lapata \shortcite{silberer2014learning} and Silberer et al. \shortcite{silberer2016visually} use stacked autoencoder to learn multimodal representations by embedding linguistic and visual inputs into a common space with the objective function of reconstructing the individual inputs. However, the above methods can only generate multimodal representations of those words that have perceptual information, thus reducing multimodal vocabulary drastically. 
	
	An empirically superior model addresses this problem by firstly predicting missing perceptual information. This includes Hill et al. \shortcite{hill2014multi} who utilize ridge regression method to learn a mapping matrix from linguistic modality to visual modality, and Collell et al. \shortcite{collell2017imagined} who employ a feed-forward neural network to learn the mapping relation between linguistic vectors and visual vectors. Applying the mapping function on linguistic representations, they obtain the predicted visual vectors for all words in linguistic vocabulary. Then they calculate multimodal representations by concatenating linguistic and predicted visual vectors. Furthermore, they find that irrelevant visual information is discarded in process of associating language to vision, which makes the predicted visual vectors outperform original visual vectors on various semantic similarity experiments.
\end{enumerate}

%Following their approach, we first obtain the predicted visual vectors. Then we propose a dynamic fusion method to combine the predicted visual representations with their corresponding linguistic representations.

According to this classification, our method falls into the second group. However, the fact that representations from different modality contribute differently to word meanings is ignored by existing models. This paper aims to solve this problem by assigning different importance weights for linguistic and perceptual representations according to different type of words, which can be seen as a weighted combination model. 

In multimodal representation models, the effectiveness of weighted combination is first emphasized by Bruni et al. \shortcite{bruni2014multimodal}, in which weights are super-parameters and the same for all words. Furthermore, Kiela et al. \shortcite{kiela2014improving} propose the Dispersion method to distinguish abstract words from concrete words, based on the observation that diversity of a word's images negatively correlates with its concreteness (in which diversity is the average cosine distance between all the visual representations of a word). Then they give zero weights to the perceptual representations of abstract words in building multimodal word representations. However, this method ignores the concreteness of each word, and can not handle words without images due to relying on visual representations for every images of a word.

\section{Proposed Method}
The problem of learning multimodal representations of a word can be formulated as $M_i = \mathcal{G}(L_i,P_i)$, where $\mathcal{G}$ is the fusion function which combines the $i^{th}$ word's linguistic representations $L_i$ with its (predicted) perceptual representations $P_i$.
In this section we describe the details of our proposed method (Figure 1): (1) build the linguistic and perceptual representations. Following most previous work, we employ visual vectors as the perceptual representations, which contain a much smaller vocabulary than linguistic vectors. (2) Learn a mapping from the linguistic to visual space. In this way, we get the predicted visual vectors for all words in linguistic vocabulary. (3) Generate multimodal representations by combining linguistic and predicted visual representations with dynamic fusion method. (4) Train the proposed model with max-margin objective function.

\begin{figure}[htb]
	\centering
	\setlength{\belowcaptionskip}{-10pt}
	\includegraphics[width=85mm,height=65mm]{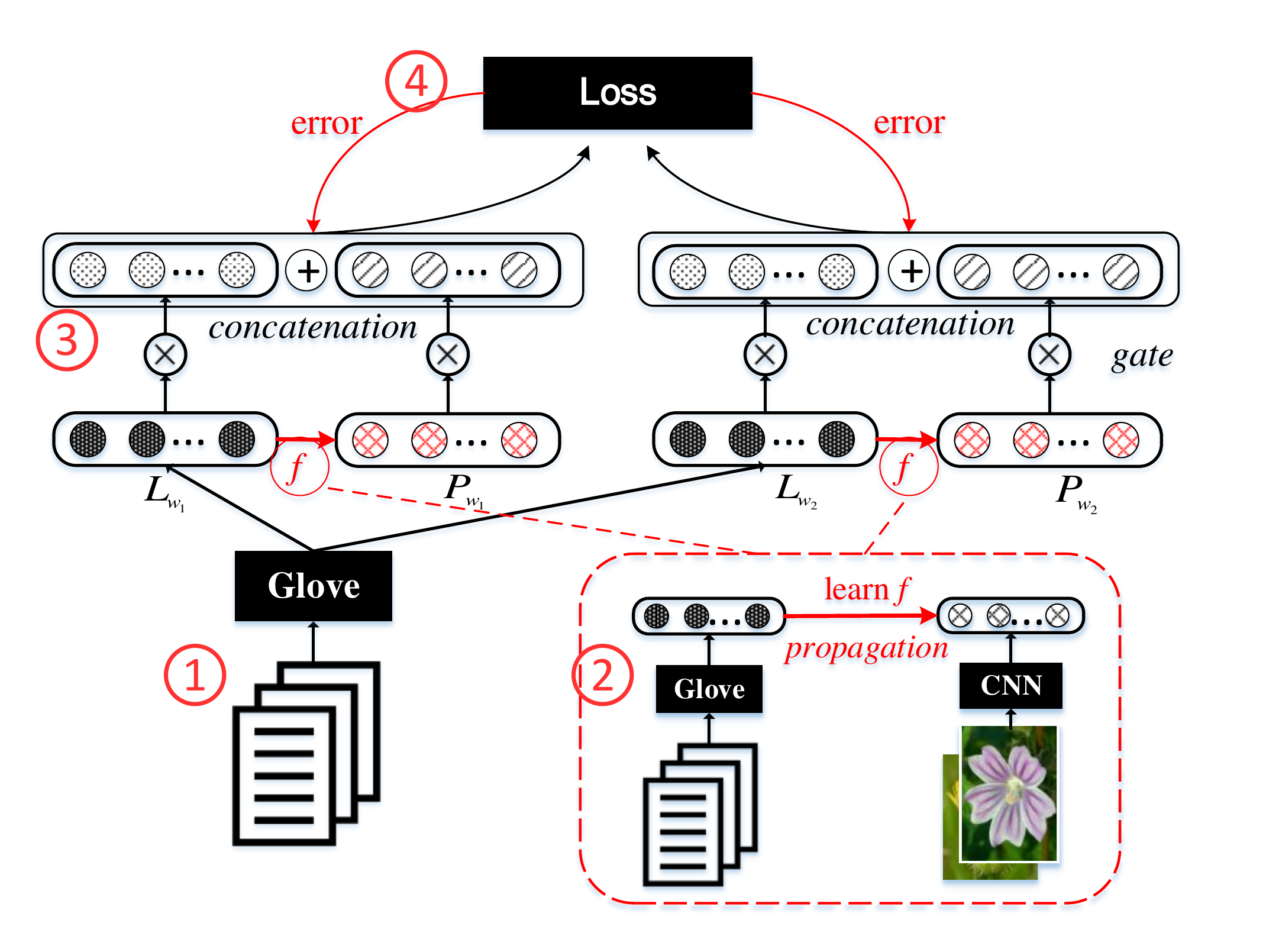}
	\caption{Overview of our model, where the four numbers correspond to four steps of our method. $L_{w_1}$ and $L_{w_2}$ are representations of one word association pair. $P_{w_1}$ and $P_{w_2}$ are the predicted visual representations from outputs of text-to-vision mapping $f$. The multimodal representation of a word is weighted concatenation of its linguistic vectors and its predicted vectors, in which weights are learnable parameters.}
\end{figure}

\subsection{Obtaining Liguistic and Visual Representations}
We employ the Glove vectors as our linguistic representations, which are trained by global word co-occurrence statistics. For visual representations, we employ image collections from ImageNet \cite{russakovsky2015imagenet}, in which each image is attached to a word and each word corresponds to multiple images. To generate visual vectors for each word, we use the forward pass of a pre-trained CNN model and extract the hidden representation of the last layer as the feature vector. Then we use averaged feature vectors of the multiple images corresponding to the same word.

\subsection{Learning to Propagate Language to Vision}
As introduced in the previous section, the words with corresponding visual images are only a small subset of the linguistic vocabulary. To obtain the visual vector for each word, we need a text-to-vision mapping function that transforms the linguistic vectors into visual ones. In this section, we introduce how to design the mapping function.

Suppose that $L \in \mathbb{R}^{m_l \times n_l}$ be the linguistic representations containing $m_l$ words, $V \in \mathbb{R}^{m_v \times n_v}$ be the visual representations of $m_v$ ($\ll m_l$) words, where $n_l$ and $n_v$ are dimensions of the linguistic and visual representations respectively. The linguistic representations of the $m_v$ visual words are denoted as $L_v \in \mathbb{R}^{m_v \times n_l}$. Our goal is to learn a mapping function from linguistic to visual space. To achieve this, we utilize ridge regression method which learns $n_v$ regression coefficients $A_j \in \mathbb{R} ^{n_l \times 1}$ that maps the linguistic representation of $L_v$ into a particular feature vector $V_j$ (the $j^{th}$ dimension of the visual representations). The objective for learning $A_j$ is then to minimize:

\begin{equation}
{||L_v A_j - V_j||_2^2}+{\lambda ||A_j||_2^2},
\end{equation}
where $\lambda$ is the regularization parameter. Finally, all $n_v$ coefficients of $A_j$ are applied together to map the $n_l$-dimensional linguistic vectors to get the $n_v$-dimensional predicted visual representations $P = LA \in \mathbb{R}^{m_l \times n_v}$.

\subsection{Generating Multimodal Representations}

\begin{figure}[htb]
	\centering
	\includegraphics[scale=0.62]{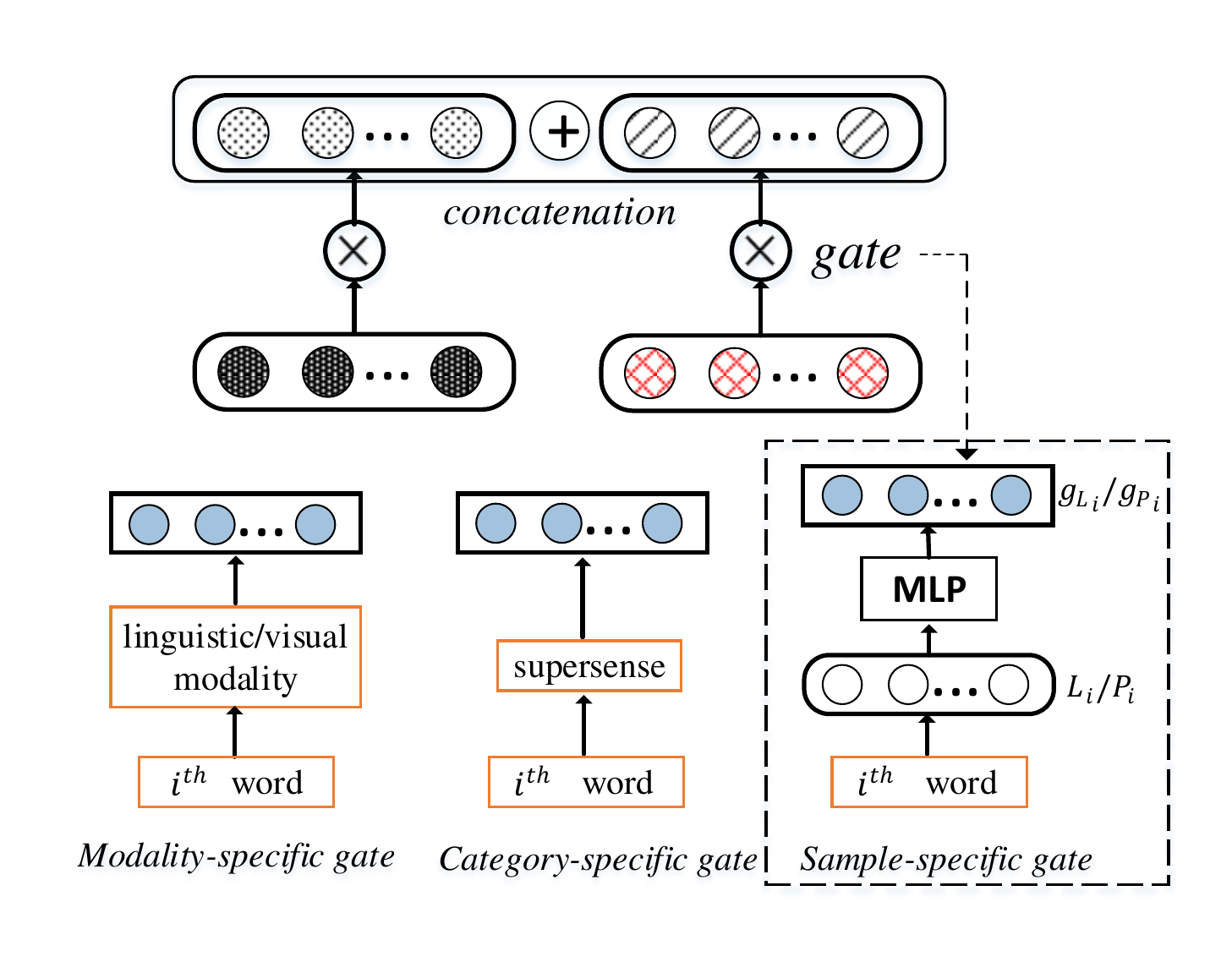}
	\caption{Illustration of three different \textit{gates}, in which each gate can be of value gate or vector gate. Take sample-specific gate (in the dashed box) as an example, it calculates importance weights $g_{L_i}$ for linguistic representations and $g_{P_i}$ for visual representations according to specific word $i$ (in which the importance weight is either a vector parameter for vector gate or a value parameter for value gate). Similarly, the category-specific gate and modality-specific gate calculate importance weights according to modality and word supersense respectively.}
\end{figure}

To build better multimodal representations, we explore three different \textit{gates} (Figure 2) to learn the importance weights of textual and predicted visual representations respectively: 

\vspace{1mm}
\noindent	\textbf{(1) Modality-specific gate}
\vspace{1mm}

\noindent	Analysis on the inner properties of linguistic and visual vectors shows that the two vectors capture some of the same properties \cite{collell2016image}, which are redundant in representing the meaning of a word. Based on this observation, we design a modality-specific gate to give a weight value or a weight vector $g_{L}$ for linguistic modality and $g_{P}$ for visual modality respectively.

\vspace{1mm}
\noindent	\textbf{(2) Categoty-specific gate}
\vspace{1mm}

\noindent	Psychological researches \cite{handjaras2016concepts} prove that human semantic representation shows clear difference between different word categories. For instance, the category of \texttt{living things} on average receives higher saliency ratings on visual properties than on \texttt{emotion} category. To model the above observation, we design a category-specific gate to give two weight values or weight vectors (i.e., $g_{L_m}$ for linguistic modality and $g_{P_m}$ for visual modality) for each word supersense\footnote{The supersense refers to 41 WordNet's supersenses  (e.g., \textit{animal, body, food, emotion, motion}), in which we tag a word with its most frequent supersense in the sense-annotated corpora: \url{https://github.com/UKPLab/acl2016-supersense-embeddings}} $m$. 

\vspace{1mm}
\noindent	\textbf{(3) Sample-specific gate}
\vspace{1mm}

\noindent	Considering that the meaning of each word has different dependencies on linguistic and visual information, we propose the sample-specific gate to assign two weight values or weight vectors (one for each modality) for each word. The weight parameters are calculated by the following feed-forward neural networks:

\begin{equation}\label{eq}
\begin{split}
g_{L_i} = tanh(W_L L_i + b_P)\\
g_{P_i} = tanh(W_P P_i + b_P),
\end{split}
\end{equation}
where $g_{L_i}$ and $g_{P_i}$ are the value gate or vector gate of the  $i^{th}$ word's linguistic representation $L_i$ and visual representation $P_i$ respectively. For the value gate, $W_L$ and $W_P$ are vector parameters with size of ${d\times 1}$, and $b_L$ and $b_P$ are value parameters. For the vector gate, the parameters $W_L$ and $W_P$ are matrices with size of ${d\times d}$, and $b_L$ and $b_P$ are vectors with size of ${d\times 1}$.

In the above dynamic fusion methods, the value gate controls the importance weights of different input representations as a whole, whereas the vector gate can adjust the importance weights of each dimension of input representations. 

Finally, we compute element-wise multiplication of the linguistic and visual representations with their corresponding gates, and concatenate the results to get the multimodal representations:

%	\begin{numcases}{M_i=}
%	& [ $g_{L} \odot L_{i}  \ ; \  g_{P} \odot P_{i}$ ] ;   \\
%	& [ $g_{L_m} \odot L_{i}  \ ; \  g_{P_m} \odot P_{i} $] \qquad \qquad \\
%	& [ $g_{L_i} \odot L_{i}  \ ; \  g_{P_i} \odot P_{i} $] \qquad \qquad
%	\end{numcases}

\begin{equation} \small
M_i =\left\{ \begin{array}{lll}
& [ g_{L} \odot L_{i}  \ ; \  g_{P} \odot P_{i} ]  & for \  \text{ M-gate}\\
& [ g_{L_m} \odot L_{i}  \ ; \  g_{P_m} \odot P_{i} ] & for \ \text{ C-gate}\\
& [ g_{L_i} \odot L_{i}  \ ; \  g_{P_i} \odot P_{i} ] & for \ \text{ S-gate}
\end{array}
\right.
\end{equation}

Where $M_i$ is the multimodal representation of the $i^{th}$ word, operator $[ v_1 ; v_2]$ denotes concatenation of vector $v_1$ and $v_2$, and $ \odot$ denotes element-wise multiplication. In the $\text{C-gate}$ model, $m$ represents the category of the $i^{th}$ word. 
%Equation (3), (4) and (5) are multimodal representations for fusion models with modality-specific gate, category-specific gate and sample-specific gate respectively.

\subsection{Training Multimodal Models}
The training data contains a set of word association pairs $(w_1, w_2)$. To learn the model parameters of different gates $p_{gates}$ (i.e., $g_{L}$ and $g_{P}$ for M-gate; $g_{L_m}$ and $g_{P_m}$ for C-gate; $W_L$, $b_L$, $W_P$ and $b_P$ for S-gate), we minimize a max-margin objective function as follows: 

\begin{equation}\label{eq} \small
\begin{split}
\sum\limits_{({w_1},{w_2}) \in W} (\max (0,1 - M_{{w_1}} \cdot M_{{w_2}} + M_{{w_1}}
\cdot M_{{n_1}}) \\+ \max (0,1 - M_{{w_1}} \cdot M_{{w_2}} + M_{{w_2}} \cdot M_{{n_2}}))
\end{split}
\end{equation}
where $M_x$ denotes the multimodal representation of word $x$ which can be calculated by equation (3), and $n_1$ and $n_2$ are randomly selected negative examples. The intuition for this objective is that we want the two association words to be more similar to each other than the negative examples.

\section{Experimental Setup}

\subsection{Datasets}
We use 300-dimensional GloVe vectors\footnote{\url{http://nlp.stanford.edu/projects/glove}} which are trained on the Common Crawl corpus consisting of 840B tokens and a vocabulary of 2.2M words. Our source of visual vectors are collected from ImageNet \cite{russakovsky2015imagenet}， which covers a total of 21,841 WordNet synsets \cite{fellbaum1998wordnet} that have 14,197,122 images. For our experiments, we delete words with fewer than 50 images or  words not in the Glove vectors, and sample at most 100 images for each word. We use a pre-trained model of VGG-net\footnote{\url{http://www.vlfeat.org/matconvnet/}} to embed visual information, resulting in 8048 vectors of 128 dimensions. 

The training dataset are selected from about 20,000 word association pairs, and each word pair is generated by at least one subject\footnote{The dataset is collected by \cite{depredicting} and can be found at: \url{https://simondedeyne.me/data}.}. We calculate the association score for each word pair (cue word + target word) as: \textit{the number of person who generated the word pair divide the total number of person who is presented with the cue word}. To select high-quality association pairs, we delete those whose score is lower than 0.2 or with words that are not in the Glove vocabulary. For better generalization ability, we delete word pairs that contain words in the testing datasets, which results in 1,494 word pairs. For the development set, we use the remaining (about 18,500) word association pairs together with their association scores. 

%Note that the development data are pairs of $<$cue word, target word, association score$>$ (in order to be closer to testing data), while training data are pairs of $<$cue word, target word$>$.

%\footnote{We randomly sample 30 word association pairs, and find that the ratio of concrete words to abstract words is about 1:3.}

\subsection{Model Settings}
Our models are implemented with Theano \cite{bergstra2010theano} and Lasagne \cite{dieleman2015lasagne}, and optimized with Adagrad \cite{duchi2011adaptive}. We test the initial learning rate over \{0.05, 0.01, 0.5, 0.1\}, set batch size to 25, and train the model for 5 epochs. We set the initial parameters in three gates to 1.0 and select the best parameters on the development set. All models are trained for 3 times and the average results are reported in Table 1. Note that we do not update word embeddings because 1) words in the training dataset are not in the testing dataset, and 2) more importantly we aim to learn generic composition rules. \textit{The data and code for training and evaluation will be released}.

\section{Experiments}
\subsection{Evaluation Tasks}
We test the baseline and proposed models on 6 standard evaluation benchmarks, covering two different tasks: (i) Semantic relatedness: Men-3000 \cite{bruni2014multimodal} and Wordrel-252 \cite{agirre2009study}; (ii) Semantic similarity: Simlex-999 \cite{hill2016simlex}, Semsim-7576 \cite{silberer2014learning}, Wordsim-203 and Simverb-3500 \cite{gerz2016simverb}. All test sets contain a list of word pairs along with their subject ratings.

We employ Spearman's method to evaluate the performance of our models. This method calculates the correlation coefficients between model predictions and subject ratings, in which the model prediction is the cosine similarity between semantic representations of two words.

\subsection{Baseline Multimodal Models}
For fair comparison, we re-implement several representative systems with our own linguistic and visual vectors. The \textbf{Concatenation (CONC) model} \cite{kiela2014learning} is simple concatenation of normalized linguistic and visual vectors. The \textbf{Ridge} \cite{hill2014multi}  and \textbf{Mapping} \cite{collell2017imagined} models first learn a mapping matrix from linguistic modality to visual modality using ridge regression method and feed-forward neural network respectively. After applying the mapping function on the linguistic representations, they obtain the predicted visual vectors for all words in linguistic vocabulary. Then they concatenate the normalized linguistic and predicted visual vectors to get multimodal representations. All above models are implemented with sklearn\footnote{\url{http://scikit-learn.org/}}. Model hyper-parameters are tuned by 5-fold cross validation (20\% of data for testing and 80\% for training) with evaluation metric of mean square error\footnote{In Ridge model, the optimal  regularization parameter is 0.6. The Mapping model is trained with SGD for maximum 100 epochs with early stopping, and the optimal learning rate is 0.001.}. The \textbf{Dispersion model} computes multimodal representations by using weighted concatenation of linguistic and visual representations, in which weights are 1 or 0 for two modalities according to whether it is concrete or abstract words. Same as Kiela et al. \shortcite{kiela2014improving}, we set the threshold (which distinguish concrete and abstract words) as the median image dispersion\footnote{We calculate image dispersion with the toolkit: \url{https://github.com/douwekiela/mmfeat}}, and give zero weights to the visual representations for abstract words before concatenation.

\subsection{Results and Discussion}

\begin{figure*}[htb]
	\centering
	\captionsetup{labelformat=empty}
	\caption{Table 1: Spearman correlations between model predictions and human ratings on six evaluation datasets. The bold scores are the best results per column and \#inst. denotes the number of word pairs. For each test, ALL correspond to the whole testing set, VIS (visual) to those word pairs for which we have both visual and linguistic vectors, and ZS (zero-shot) denotes word pairs for which we have only linguistic vectors. We divide all models into four groups: (1) previous multimodal models in which results are reprinted from \cite{collell2017imagined}. (2) Uni-modal models with linguistic and visual inputs respectively. (3) Our re-implementation of baseline multimodal models. (4) Our proposed multimodal models with the dynamic fusion method, where \texttt{M,C,S} denote modality-specific, category-specific and sample-specific respectively. \texttt{gate-val} and \texttt{gate-vec} denote value gate and vector gate respectively.}
	\includegraphics[scale=0.88]{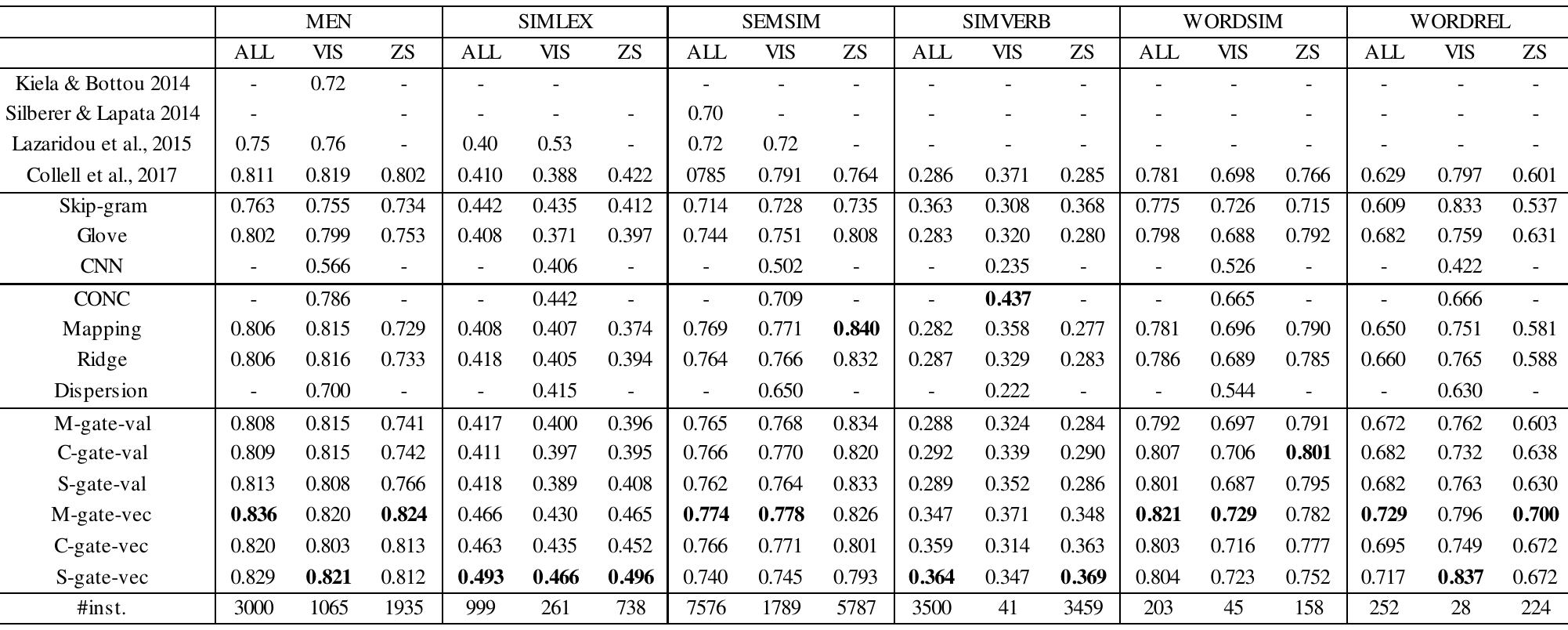}
	\captionsetup{labelformat=empty}
	
\end{figure*}

As shown in Table 1, our proposed multimodal models clearly outperform baseline unimodal and multimodal models (in group 2 and 3). We use Wilcoxon signed-rank test to check if significant difference exists between two models. Results show that our multimodal models with vector gates perform significantly better ($p < 0.05$) than all baseline models, while the multimodal models with value gates do no show significant difference over Ridge model.

\vspace{1mm}
\noindent	\textbf{Overall performance}
\vspace{1mm}
Our multimodal models with vector gate (i.e., M-gate-vec, C-gate-vec, S-gate-vec) achieve better performance than Ridge in VIS (visual, the testing data that have associated visual vectors) and ZS (zero-shot, the testing data that do not have associated visual vectors) region. In other words, our models improve Ridge on words with (mostly concrete words) or without visual information (more abstract words). This suggests that the dynamic fusion methods can dynamically fuse different modality inputs. The good results in ZS region also indicate that our models have good generalization capacity. Therefore, our multimodal representations with vector gates clearly accomplish one of their foremost goals, namely to improve the multimodal representations for all types of words. 

\vspace{1mm}
\noindent	\textbf{Unimodal baselines}
\vspace{1mm} 
On the linguistic side, we additionally test Skip-gram model. Comparing unimodal models (in group 2), we can see that Glove outperforms Skip-gram on four datasets while Skip-gram takes superiority on the other two datasets, indicating that these two text-based models may encode different types of information. The CNN model, which learns representations from visual modality, gets worse performance than Glove and Skip-gram. 

\vspace{1mm}
\noindent	\textbf{Multimodal baselines}
\vspace{1mm}
The CONC model that combines Glove vectors and CNN visual vectors, performs worse than Glove on four out of six datasets, suggesting that simple concatenation might be suboptimal. The Mapping and Ridge models, which combine Glove vectors and predicted visual vectors, improve over Glove on five out of six datasets in both ALL and VIS regions. This indicates that the predicted visual vectors contain richer information than purely visual representations and are more helpful in building multimodal models. In ZS region, multimodal models of Mapping and Ridge only significantly outperform Glove on the \textit{SEMSIM} dataset.
 
\vspace{1mm}
\noindent	\textbf{Our multimodal models}
\vspace{1mm}
Among our proposed models, the multimodal models with vector gate are clearly better than the ones with value gate (i.e., M-gate-val, C-gate-val, S-gate-val). This indicates that combining representations from different modalities is more complex than weighted concatenation, and thus needing deep fusion methods that can selectively combine the inside elements of different representations. Another observation is that the multimodal model with category-specific vector gate is not as effective as other two models with vector gates. This is possibly due to that the tagging process of word supersense introduces some errors.

\subsection{Model Analysis}

%\textbf{Effects of training data size. }
\subsubsection{Effects of training data size}

To investigate the effects of training data size, we conduct experiments with less training data. As can be seen in Figure 3, decreasing the number of training data clearly harms the performance of models with vector gate, but leads to no obvious difference for the models with value gate. Additionaly, we observe that the models with vector gate can obtain a quite good result with 40\% of data (600 training pairs), indicating that our dynamic fusion methods can be successfully trained with a relative small training set.

\begin{figure}
	\centering
	\includegraphics[scale=0.6]{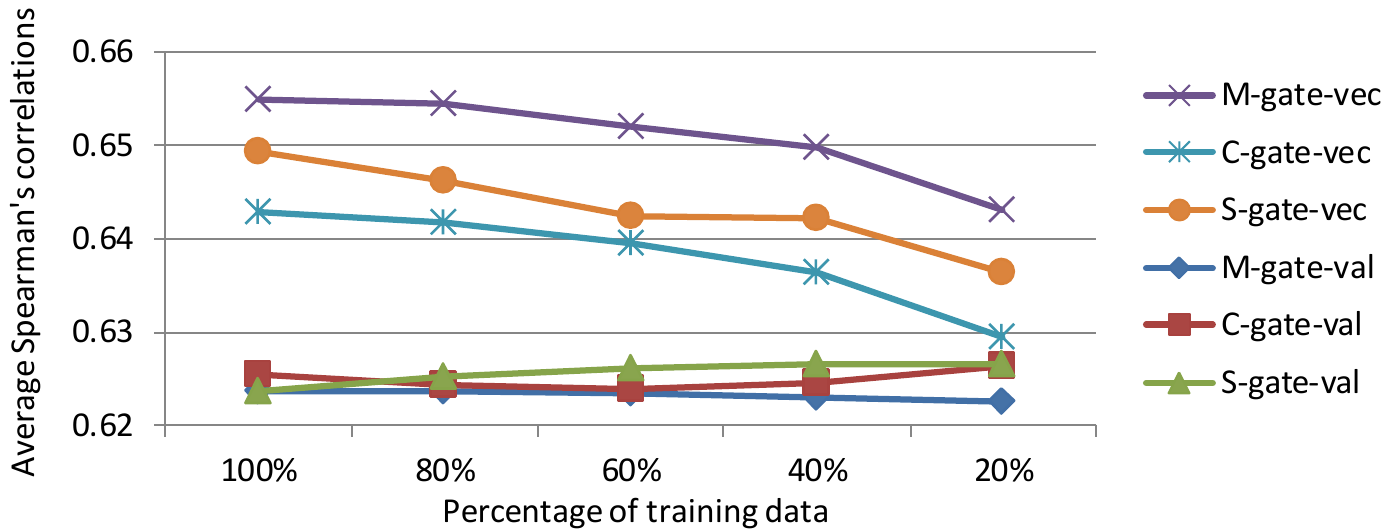}
	\captionsetup{labelformat=empty}
	\caption{Figure 3: Effects of training data size on the model performance, which are evaluated by averaged Spearman's correlations on all evaluation datasets.}
\end{figure}

%\noindent \textbf{Effects of different gates. }
\subsubsection{Effects of different gates}
To inspect whether the proposed models meet our expectation, i.e., assigning different weights to linguistic and visual representations for concrete words and abstract words respectively, we conduct a quantitative analysis using a set of concrete and abstract words. Specifically, we utilize the University of South Florida dataset (USF)\footnote{The dataset can be download at: \url{http://web.usf.edu/FreeAssociation/}}, which includes concreteness ratings for over 6,000 words collected from thousands of participants\footnote{Examples of word and its concreteness are: \textit{(tree, 7), (eye, 6.28), (wind, 5.4), (dark, 4.68), (work, 3.88), (effort, 2.22), (hope, 1.18)}.}. To extract a set of abstract and concrete words, we first select words those appear in both USF dataset and the linguistic vocabulary, and order these words according to their concreteness ratings. Next we sample at random from the first and fourth quartiles, in which we get 796 concrete words and 788 abstract words. Then we examine the weights assigned to different modalities on abstract and concrete words respectively. In the following, we separately describe the results of our proposed dynamic fusion methods with different gating mechanism.

\textbf{M-gate-val} obtains a weight value of 1.089 for linguistic modality, and 0.911 for visual modality.

\textbf{C-gate-val} learns one weight value for each supersense category in linguistic and visual modality respectively. The five categories with highest weight ratio of linguistic to visual are \texttt{Attribute, Location, Cognition, Quantity}, and \texttt{State}. The five categories with lowest weight ratio of linguistic to visual are \texttt{Animal, Object, Motion, Shape}, and \texttt{Plant}\footnote{Example words in these categories: Attribute (\textit{age, power, strength}), Location (\textit{area, west, south}), Cognition (\textit{thought, believed, known}),  Quantity (\textit{miles, meters, number}), State (\textit{condition, problems, health}), Animal(\textit{cow, frogs, birds}), Object (\textit{river, valley, lake}), Motion (\textit{follow, travel, enter}), Shape (\textit{concave, angle, lines}), and Plant (\textit{fruit, flowers, vines}).}.

\textbf{S-gate-val} calculates one weight value for each word in linguistic and visual modality respectively. We then compute the average weight ratio of linguistic to visual modality respectively on the set of abstract and concrete words. As a result, we get 1.965:1 for concrete words, and 2.203:1 for abstract words. Moreover, the five words with the highest ratio of linguistic to visual are: \texttt{much, seem, curious, sense, mind}, whereas the five words with lowest ratio are: \texttt{wharf, walkway, married, beverage, tower}. In addition, we test the Spearman correlation between word concreteness and weight ratio of linguistic to visual modality for all these words, which results a correlation score of 0.614. 

\textbf{M-gate-vec} assigns one vector for each modality. To inspect the importance weight of linguistic and visual modality, we calculate the $l_2$-norm of the two vectors. Finally, we get a value of 1.186 for linguistic modality and 0.045 for visual modality.

\textbf{C-gate-vec} learns one vector for each supersense category in linguistic and visual modality respectively. We then calculate the $l_2$-norm of these vectors and the weight ratio of linguistic to visual modality in each category. The five categories with highest ratio are \texttt{Attribute, Cognition, State, Social}, and \texttt{Change}. The five categories with lowest ratio are \texttt{Shape, Object, Creation, Motion}, and \texttt{Plant}\footnote{Example words in these categories: Social (\textit{succeed, encourage, join}), Change (\textit{lengthen, simplify, diminishing}), Creation (\textit{publish, design, fix}), Motion (\textit{drop, swim, roll}).}.

\textbf{S-gate-vec} calculates one vector for each word in linguistic and visual modality respectively. We then compute the averaged $l_2$-norm weight ratio of linguistic to visual modality on the set of abstract and concrete words respectively. As a result, we get 2.975:1 for concrete words, and 3.714:1 for abstract words. Moreover, the five words with the highest ratio are: \texttt{really, think, seriously, reason, believe}, and the five words with lowest ratio are: \texttt{volcano, palace, salad, shackle, tomato}. Furthermore, we test the Spearman correlation between word concreteness and weight ratio of linguistic to visual modality for all these words, in which we get a correlation score of 0.458. 

From the above results, we observe that (1) the proposed models can successfully assign different weights to linguistic and visual modalities, and the learned weights show clear difference between concrete and abstract words. (2) For models with modality-specific gates (M-gate-val, M-gate-vec), representations of linguistic modality always achieve higher weights, which indicates that linguistic vectors are more important in building multimodal representations on the whole. (3) As for models with category-specific gates (C-gate-val, C-gate-vec), the categories which contain mostly abstract words achieve higher weight ratio of linguistic to visual modality, which means that the linguistic modality is more important to abstract words. (4) In models with sample-specific gates (S-gate-val, S-gate-vec), abstract words achieve higher weight ratio of linguistic to visual modality. Moreover, the learned weight-ratio shows high correlation with word concreteness, indicating that the proposed model can assist in related psycholinguistic experiments.

\section{Conclusions and Future Work}
Motivated by the fact that different semantic word representations require information from different modality inputs, in this paper we propose three simple but effective fusion methods for learning multimodal word representations. Experimental evaluations show that our proposed models achieve substantial gains in accuracy on all six benchmarks. Qualitative analyses further evidence that the proposed methods can dynamically fuse representations from different modalities according to different types of words.

Future work includes exploring better semantic word representations by combining information from other modality inputs like audition and olfaction. Moreover, the visual representations can be enhanced by utilizing more fine-grain semantic understanding of an image, which can be achieved with measures like image segmentation. We believe that one of the promising directions is learning from human semantic representation to build a more cognitive-inspired multimodal model.

\bibliography{aaai}
\bibliographystyle{aaai}

\end{document}